\documentclass[10pt,twocolumn,letterpaper]{article}

\usepackage[pagenumbers]{cvpr} 

\usepackage{graphicx}
\usepackage{amsmath}
\usepackage{amssymb}
\usepackage{booktabs}
\usepackage[table,xcdraw,dvipsnames]{xcolor}

\usepackage{multirow}
\usepackage[table,xcdraw]{xcolor}

\usepackage[pagebackref,breaklinks,colorlinks]{hyperref}

\usepackage[capitalize]{cleveref}
\crefname{section}{Sec.}{Secs.}
\Crefname{section}{Section}{Sections}
\Crefname{table}{Table}{Tables}
\crefname{table}{Tab.}{Tabs.}

\newcommand{\ourmethod}{{HEAT}\xspace}
\newcommand{\ourmethodtable}{{HEAT(Ours)}\xspace}

\newcommand{\first}[1]{\textcolor{cyan}{#1}}
\newcommand{\second}[1]{\textcolor{orange}{#1}}

\usepackage{amssymb}
\usepackage{amsmath,amsfonts}
\usepackage{amsopn,bm}

\usepackage{nicefrac}       

\usepackage{epsfig}
\usepackage{graphicx}
\usepackage{float}
\usepackage{wrapfig}
\usepackage[ruled,vlined]{algorithm2e}

\usepackage{bm,xspace}
\usepackage{comment}
\usepackage{verbatim}
\usepackage{multirow}
\usepackage{makecell}
\usepackage{array}
\usepackage{balance}
\usepackage{url}
\usepackage{booktabs}
\usepackage{etoolbox,siunitx}
\usepackage{calc}
\usepackage{pifont,hologo}
 
\usepackage{nicefrac}

\usepackage{arydshln}
\usepackage[utf8]{inputenc} 
\usepackage[T1]{fontenc}    
\usepackage{url}            
\usepackage{amsfonts}       
\usepackage{nicefrac}       
\usepackage{microtype}      
\usepackage[american]{babel}
\usepackage{enumitem}
\usepackage{siunitx}

\usepackage{enumitem}

\usepackage[super]{nth}
\usepackage{nicefrac}
\robustify\bfseries

\usepackage{dsfont}

\usepackage{listings}
\usepackage{xcolor}
\definecolor{codegreen}{rgb}{0,0.6,0}
\definecolor{codegray}{rgb}{0.5,0.5,0.5}
\definecolor{codepurple}{rgb}{0.58,0,0.82}
\definecolor{backcolour}{rgb}{1.0,1.0,1.0}

\lstdefinestyle{mystyle}{
    commentstyle=\color{codegreen},
    keywordstyle=\color{magenta},
    numberstyle=\tiny\color{codegray},
    stringstyle=\color{codepurple},
    basicstyle=\ttfamily\footnotesize,
    breakatwhitespace=false,         
    breaklines=true,                 
    captionpos=b,                    
    keepspaces=true,                 
    numbersep=0pt,                  
    showspaces=false,                
    showstringspaces=false,
    showtabs=false,                  
    tabsize=2,
    linewidth=.99\textwidth,
    xleftmargin=0.01cm
}
\usepackage{mathtools}

\newcommand{\cmark}{\text{\ding{51}}}

\newcommand{\ProbOpr}[1]{\mathbb{#1}}

\newcommand{\expect}[2]{%
\ifthenelse{\equal{#2}{}}{\ProbOpr{E}_{#1}}
{\ifthenelse{\equal{#1}{}}{\ProbOpr{E}\left[#2\right]}{\ProbOpr{E}_{#1}\left[#2\right]}}} 
\newcommand{\var}[2]{%
\ifthenelse{\equal{#2}{}}{\ProbOpr{VAR}_{#1}}
{\ifthenelse{\equal{#1}{}}{\ProbOpr{VAR}\left[#2\right]}{\ProbOpr{VAR}_{#1}\left[#2\right]}}} 

\makeatletter
\DeclareRobustCommand\onedot{\futurelet\@let@token\@onedot}
\def\@onedot{\ifx\@let@token.\else.\null\fi\xspace}

\def\eg{\emph{e.g}\onedot} 
\def\ie{\emph{i.e}\onedot}

\def\etal{\emph{et al}\onedot}
\makeatother

\newcommand{\eat}[1]{{}}

\newcommand\mypara[1]{\vspace{1mm}\noindent\textbf{#1}}

\usepackage{tabu}

\definecolor{Gray}{gray}{0.5}

\newlength\savewidth

\makeatletter\renewcommand\paragraph{\@startsection{paragraph}{4}{\z@}
  {.5em \@plus1ex \@minus.2ex}{-.5em}{\normalfont\normalsize\bfseries}}\makeatother

\newcolumntype{x}[1]{>{\centering\arraybackslash}p{#1pt}}
\newcolumntype{y}[1]{>{\raggedright\arraybackslash}p{#1pt}}
\newcolumntype{z}[1]{>{\raggedleft\arraybackslash}p{#1pt}}

\definecolor{Highlight}{HTML}{39b54a}  

\def\cvprPaperID{} 
\def\confName{}
\def\confYear{}

\begin{document}

\title{HEAT: Holistic Edge Attention Transformer for Structured  Reconstruction}

\author{Jiacheng Chen$^{1}$ \qquad Yiming Qian$^{2}$ \qquad Yasutaka Furukawa$^{1}$
\\[6pt]
$^1$Simon Fraser University \qquad $^2$University of Manitoba
}

\twocolumn[{
 \maketitle
 \vspace{-2em}
 \centerline{
 \includegraphics[width=\textwidth]{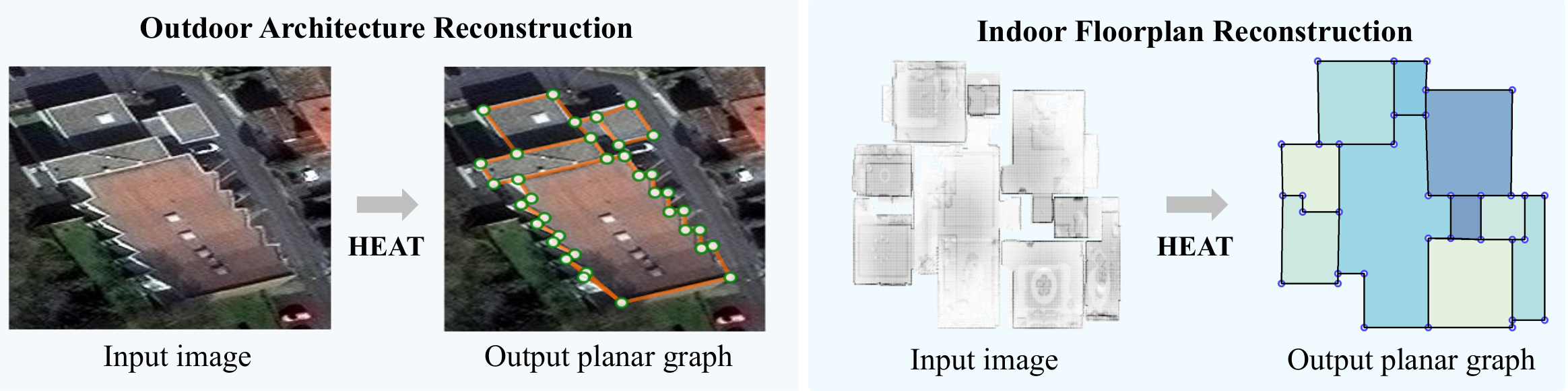} 
}
\captionof{figure}{HEAT takes a 2D raster image as an input and reconstructs a planar graph by an end-to-end transformer based neural architecture, for example, from a satellite image to an outdoor building architecture, or from a point-density image to an indoor floorplan.}
\label{fig:teaser}
\vspace{1em}
}]

\begin{abstract}
This paper presents a novel attention-based neural network for structured reconstruction, which takes a 2D raster image as an input and reconstructs a planar graph depicting an underlying geometric structure. The approach detects corners and classifies edge candidates between corners in an end-to-end manner. Our contribution is a holistic edge classification architecture, which
1) initializes the feature of an edge candidate by a trigonometric positional encoding of its end-points; 2) fuses image feature to each edge candidate by deformable attention; 3) employs two weight-sharing Transformer decoders to learn holistic structural patterns over the graph edge candidates; and 4) is trained with a masked learning strategy. The corner detector is a variant of the edge classification architecture, adapted to operate on pixels as corner candidates. We conduct experiments on two structured reconstruction tasks: outdoor building architecture and indoor floorplan planar graph reconstruction. Extensive qualitative and quantitative evaluations demonstrate the superiority of our approach over the state of the art. Code and pre-trained models are available at \url{https://heat-structured-reconstruction.github.io/}
\end{abstract}

\section{Introduction}
\label{sec:intro}
Human vision has a remarkable capability in holistic structure reasoning. Looking at the building in Figure~\ref{fig:teaser}, we can effortlessly identify the structural primitives (\eg, building corners and edges) and their relationships. 
A fundamental challenge in computer vision research is to acquire such human-level perceptual capability and ultimately reconstruct holistic geometric structures from an image, which would have tremendous impacts on broader domains such as visual effects, construction, manufacturing, and robotics.

Ever since the emergence of deep neural networks (DNNs), geometry reconstruction has seen a breakthrough in low-level primitive detection tasks (\eg, corners and edges) by training DNNs on large annotated datasets~\cite{Xu2021LETR,Xue2020HAWP,Zheng2020Structured3DAL,Zhou2019wireframe3D}.
However, holistic structure reasoning (\eg, graph inference from corner candidates) is still a challenge for end-to-end neural architectures. The performance is far below that of human-vision~\cite{Zhang2020ConvMPN} and inferior to classical optimization or search methods~\cite{Nauata2020VectorizingWB,Chen2019FloorSPIC,Stekovic2021MonteFloorEM}.

This paper seeks to push the frontier of end-to-end neural architecture for structured reconstruction, in particular, we focus on inferring an outdoor building architecture or an indoor floorplan as a 2D planar graph from raster sensor data.
Our approach resembles state-of-the-art Transformer based architectures DETR/LETR for object/edge detection~\cite{Carion2020DETR,Xu2021LETR}. They extract image information by a ConvNet and pass to the ``dummy query nodes'' in a Transformer decoder by cross-attention.
Our experiments reveal that they do not effectively learn structural regularities of target objects/edges.

Our idea is similar and simple, yet yields much more powerful holistic structural reasoning. 
For edge detection, instead of creating ``dummy query nodes'' as placeholders for regressing answers, we 1) create a node for every edge candidate and initialize the feature by a trigonometric positional encoding of its end-points~\cite{Vaswani2017Transformer}; 2) fuse multi-scale image features from ConvNet backbone to each edge candidate by an adapted deformable attention mechanism~\cite{Zhu2021DeformableDETR} and filter out edge candidates; 3) learn structural patterns of edges by two weight-sharing Transformer decoders, one of which only sees the positional encoding (w/o image features) to enhance geometry learning; and 4) employ a masked learning strategy~\cite{Devlin2019BERT} for end-to-end training and iterative inference. A simple adaptation of the edge classification architecture is our corner detector, which further improves performance.

We have evaluated the proposed approach on two structured reconstruction benchmarks: outdoor building reconstruction from a satellite image and indoor floorplan reconstruction from a point-cloud density image.
The qualitative and quantitative evaluations demonstrate that our approach outperforms all the competing methods for the outdoor reconstruction task. For the indoor reconstruction task, our approach outperforms all fully-neural methods and is on par with domain specific solutions~\cite{Stekovic2021MonteFloorEM,Chen2019FloorSPIC}, which are more than 1000 times slower with the use of optimization and search.

\section{Related Work}
\label{sec:related}
We review structured reconstruction algorithms in three groups: classical techniques, hybrid approaches incorporating deep learning, and end-to-end neural systems.

\subsection{Classical techniques}
Structured geometry reconstruction has been an active area of research in computer vision, turning raster sensor data into vectorized geometries including wireframes, planes~\cite{furukawa2009manhattan,silberman2012indoor}, room layouts~\cite{delage2006dynamic,hedau2009recovering}, floorplans~\cite{cabral2014piecewise}, and polygonal loops \cite{gimenez2015reconstruction}. Traditional methods rely on low-level image processing techniques such as Hough transform \cite{llados1997system,adan20113d} or superpixel segmentation \cite{qin2018accurate}. More sophisticated solvers are also proposed, for example, graphical model inference based on graph-cuts for planar reconstruction \cite{furukawa2009manhattan,silberman2012indoor}, dynamic programming for floorplan recovery \cite{cabral2014piecewise}, and Bayesian network for room layout estimation \cite{delage2006dynamic}. These methods involve a plethora of heuristics or parameters to tweak by hand.

\subsection{Hybrid approaches}
With the development of neural networks, deep learning has become a \emph{de facto} machinery for vector-geometry reconstruction. Many state-of-the-art systems adopt a two-stage pipeline, where neural networks first detect low-level primitives (\eg, corners, edges, region segments) then optimization techniques assemble them into the final models~\cite{liu2017raster,zou2018layoutnet,qian2020learning}.
Nauata \etal~\cite{Nauata2020VectorizingWB} and FloorSP \cite{Chen2019FloorSPIC} rely on Mask R-CNN \cite{He2020MaskR} for primitive detection and apply optimization techniques (\eg, integer programming) to reconstruct outdoor building and indoor floorplan as planar graphs, respectively. MonteFloor~\cite{Stekovic2021MonteFloorEM} takes similar detection framework while resorting to Monte Carlo Tree Search for reconstructing the graph structure.
Albeit effective, optimization/search requires domain-specific algorithm design by hand and is a few orders of magnitude slower at test time.
A recent approach by Zhang \etal iterates exploration and classification steps to search for a better solution~\cite{Zhang2021ExploreClassify}, while our approach is much faster and performs better.

\subsection{End-to-end neural systems}
End-to-end neural architectures require less hand-engineering and achieve fast inference. For the wireframe parsing task~\cite{Huang2018wireframe2D},
L-CNN \cite{Zhou2019EndtoEndWP} adapts a likelihood prediction convolutional network (ConvNet) for junction detection, followed by an edge verification network to classify each line candidate.  
PPGNet \cite{zhang2019ppgnet} and HAWP \cite{Xue2020HAWP} also use the two-stage framework as L-CNN while proposing more advanced model designs. Zhou \etal~\cite{Zhou2019wireframe3D} extends the wireframe task to 3D by estimating depths and vanishing points together with the geometry primitives. These techniques handle edge candidates independently, while our approach jointly infers an overall structure by learning holistic structural patterns.
ConvMPN is a special graph neural network designed for planar graph reconstruction~\cite{Zhang2020ConvMPN}, while our attention-based architecture yields much better results.

The recent success of Transformer-based object detector DETR \cite{Carion2020DETR} has also been extended to wireframe parsing by LETR \cite{Xu2021LETR}. DETR/LETR utilizes ``dummy nodes'' as placeholders for storing detection answers and refrain from heuristic-based steps like non-maximum suppression. Our approach designs decoders and training strategies over all edge candidates instead of dummy nodes, and demonstrates more effective holistic structural reasoning.

\begin{figure*}[ht]
\centering
\includegraphics[width=\linewidth]{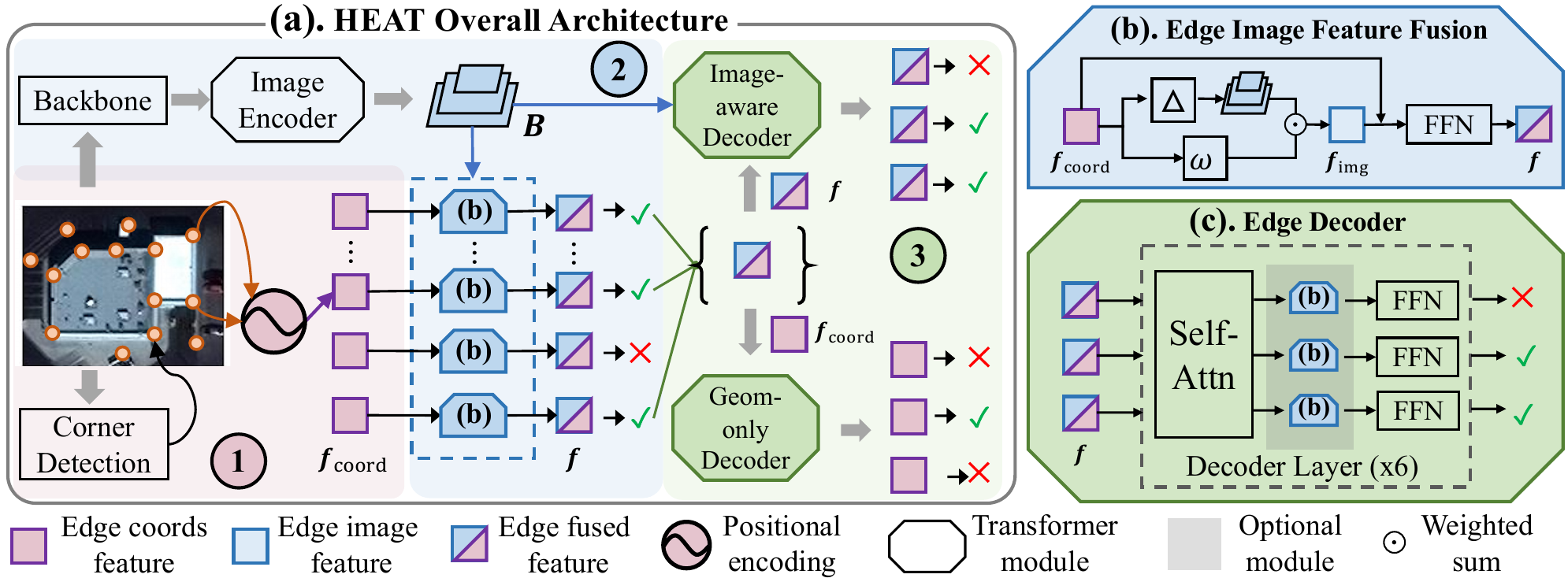}
\vspace{-1.5em}
\caption{\textbf{(a)}. The overall architecture of \ourmethod, which consists of three steps: 1) edge node initialization; 2) edge image feature fusion and edge filtering; and 3) holistic structural reasoning with two weight-sharing Transformer decoders.  \textbf{(b)}. The image feature fusion module for edge nodes. \textbf{(c)}. The edge Transformer decoder. For the geometry-only (geom-only) decoder, $\bm{f}$ is replaced by $\bm{f}_{\text{coord}}$ and the image feature fusion module (gray part) is discarded. The add-norm layers in (b) and (c) are omitted for simplicity.
} 
\label{fig:method}
\end{figure*}
\section{Preliminaries}
\label{sec:problem}
The paper borrows two structured architecture reconstruction tasks in the literature to demonstrate our approach, which are to infer a planar graph (\ie, corners and edges) depicting an architectural structure given a 2D raster image.

\mypara{Outdoor architecture reconstruction} is borrowed from the work by Nauata \etal~\cite{Nauata2020VectorizingWB}. A cropped satellite image is the input, containing one building from either Paris, Las Vegas, or Atlanta. An output planar graph depicts a roof structure, where each flat roof component is bounded by edges (See Fig.~\ref{fig:teaser}). The images come from the SpaceNet Challenge~\cite{Etten2018SpaceNetAR} and are distributed under the CC BY-SA 4.0 license. The dataset contains 1601, 50, and 350 samples for training, validation, and test, respectively. The precision/recall/F-1 scores of the corner/edge/region primitives are the metrics as in their work~\cite{Nauata2020VectorizingWB}. This task is more challenging than the following floorplan task, because satellite images suffer from perspective distortions and the Manhattan assumption does not hold.
Furthermore, a roof component cannot be reliably extracted by an instance segmentation technique. In the floorplan task, rooms can be detected easily which are utilized in state-of-the-art floorplan reconstruction techniques~\cite{Chen2019FloorSPIC,Stekovic2021MonteFloorEM}. An input image crop is of resolution 256 $\times$256 or 512$\times$512. The average/maximum numbers of corners and edges in the graphs are 12.6/93 and 14.2/101, respectively.

\mypara{Floorplan reconstruction} 
recovers the floorplan of an indoor scene as a planar graph from a point-cloud density image in the top view~\cite{Chen2019FloorSPIC,Stekovic2021MonteFloorEM}. In particular, we take the Structured3D dataset~\cite{Zheng2020Structured3DAL},
convert the registered multi-view RGBD panorama images of each scene to a point cloud, project the 3D points to an XY image plane, and generate a 256$\times$256 density image where each pixel is the number of projected points after normalization (See Fig.~\ref{fig:teaser}). The benchmark contains a total of 3500 scenes (3000/250/250 for training/validation/test) with a diverse set of house floorplans covering both Manhattan and non-Manhattan layouts. The average/maximum numbers of corners and edges across all floorplan graphs are 22.0/52 and 27.6/74, respectively. Following the recent work MonteFloor~\cite{Stekovic2021MonteFloorEM}, we use the same precision/recall of room/corner/angle as the evaluation metrics. Note that the metrics require a set of closed polygons for evaluation. Therefore, edges that are not part of a closed polygon are discarded before computing the metric.

\section{Holistic Edge Attention Transformer}
\label{sec:method}

Holistic edge attention transformer (HEAT) consists of three steps: 1) edge node initialization; 2) image feature fusion and edge filtering; and 3) holistic edge self-attention for structural reasoning (See Fig.~\ref{fig:method}). The section explains the three steps, followed by the description of 4) the training/inference schemes, and 5) the corner detector, which is an adaptation of HEAT. 
Input images are either 256$\times$256 or 512$\times$512, where we explain the architecture parameters in the former case. In the latter case, the spatial resolutions of the ConvNet features simply become double, while the rest of the architecture remains the same. We refer the full network specification to the supplementary.

\subsection{Edge node initialization}
\label{subsec:edge-init}

We start with detecting a set of corner candidates from the input image by a corner detector (See Sect.~\ref{section:corner_detector}).
Each pair of corners is an edge candidate and becomes a Transformer node, whose feature $\bm{f}_{\text{coord}}$ is initialized by the 256-dimensional trigonometric positional encoding~\cite{Vaswani2017Transformer}:
\begin{eqnarray*}
    \bm{f}_{\text{coord}} &=& \bm{M}_{\text{coord}} \left[\bm{\gamma}(e_1^x),\  \bm{\gamma}(e_1^y),\ \bm{\gamma}(e_2^x),\  \bm{\gamma}(e_2^y)\right], \\
    \bm{\gamma}(t) &=& \left[ \sin(w_0 t), \cos(w_0 t), \cdots \sin(w_{31} t), \cos(w_{31} t) \right],\\
w_i &=& 
\left(1/10000\right)^{2i/32} \ \ (i = 0, 1, \cdots 31).
\notag
\end{eqnarray*}
$e_1$ and $e_2$ are the two corners. $e^x_1$ (resp. $e^y_1$) denotes the x (resp. y) coordinate of $e_1$. $\bm{M}_{\text{coord}}$ is a $256\times 256$ learnable matrix for linear mapping. 
The function $\gamma$ encodes ordinal priors including relative distance between coordinates.

\subsection{Image feature fusion and edge filtering}
\label{subsec:fuse-image}

We inject image features into each edge node by adapting a deformable attention technique, originally developed for object detection in deformable-DETR~\cite{Zhu2021DeformableDETR}. The technique enables adaptive attention over the image features, while existing edge-feature initialization techniques sample pixels uniformly along an edge~\cite{Zhou2019EndtoEndWP,Xue2020HAWP}.

We use a ResNet~\cite{He2016Resnet} backbone and a Transformer encoder borrowed from deformable-DETR~\cite{Zhu2021DeformableDETR} to build a 3-level image feature pyramid, whose shapes are 64x64x256, 32x32x256, and 16x16x256, respectively.
For each level $l (= 1, \cdots 3)$ of the feature pyramid, we use $\bm{f}_{\text{coord}}$ to generate sampling locations around an edge, as well as their attention-weights for aggregation:
\begin{align}
    \bm{\Delta}^l &= \bm{M}_{loc}^l \bm{f}_{\text{coord}},\\
    \bm{w}^l &= \text{softmax}\left(\bm{M}_{\text{agg}}^l \bm{f}_{\text{coord}} \right).
\end{align}
$\bm{M}_{loc}^l$ and $\bm{M}_{\text{agg}}^l$ are learnable weights for the feature level $l$. $\bm{\Delta}^l$ contains four 2D sampling offsets with respect to the edge center for the feature level $l$, and $\bm{w}^l$ is the corresponding attention weights after the softmax over all the levels and samples. 
The image feature $\bm{f}_{\text{img}}$ at each edge is given by
\begin{align}
    \bm{f}_{\text{img}} = \sum_{l=1}^{3} \sum_{i=1}^{4} \bm{w}^l(i) \left[\bm{M}^{l}_{\text{img}}\ \bm{B}^l \left(\frac{e_1 + e_2}{2^{l+2}} +\frac{\bm{\Delta}^l(i)}{2^{l+1}} \right) \right]
\end{align}
$\bm{f}_{\text{img}}$ is a 256-dimensional vector.
$\bm{M}^l_{\text{img}}$ is a learnable weight matrix for level $l$.
$\bm{B}^l$ denotes the feature map at level $l$ of its feature pyramid.
Note that we also employ 8-way multi-head attention strategy.
Finally, we obtain a fused feature $\bm{f}$ by a standard add-norm layer and a feed forward network (FFN) as in the original Transformer~\cite{Vaswani2017Transformer}:
\begin{equation}
    \bm{f} = \text{FFN}\left(\text{Add\&Norm}\left(\bm{f}_{\text{img}}, \bm{f}_{\text{coord}}\right)\right). \label{eq:f_fusion}
\end{equation}
We illustrate the image feature fusion module in Figure~\ref{fig:method}b.

\mypara{Filtering edge candidates:} 
Transformer is memory intensive and we filter out bad candidates by passing $\bm{f}$ to a 2-layer MLP followed by a sigmoid function and computing a confidence score. We keep the top-$K$ candidates, where $K$ is three times the number of corner candidates. This filtering often takes a few thousand candidates and cuts down to a few hundred.
This module is trained with the rest of the network with a binary cross entropy (BCE) loss.
We have four BCE losses in total and refer to Sect.~\ref{sec:exp} for the balancing weights.

\subsection{Holistic edge decoders}
\ourmethod adopts two weight-sharing Transformer~\cite{Vaswani2017Transformer} decoders to classify each edge candidate to be correct or not.

\mypara{Image-aware decoder:} The first decoder takes the fused feature $\bm{f}$ for each edge candidate as a node. The network contains six layers of (self-attention, edge image feature fusion module, and feed forward network), with 8-way multi-head attention (See Fig.~\ref{fig:method}c).  The BCE is the loss function.

\mypara{Geometry-only decoder:} The second decoder has exactly the same architecture and shares the weights, while we disable the use of image information. First, we pass the coordinate feature $\bm{f}_{\text{coord}}$ as the initialization, which does not have image information. Second, this decoder does not have the image feature fusion module (gray parts in Fig.~\ref{fig:method}c). The decoder is forced to solve the task solely with geometry information, enhancing holistic geometric reasoning and the performance of the image-aware decoder through weight-sharing. This decoder is trained with BCE loss, while only the image-aware decoder is used for test time results.

\begin{figure}[!t]
    \centering
    \includegraphics[width=\linewidth]{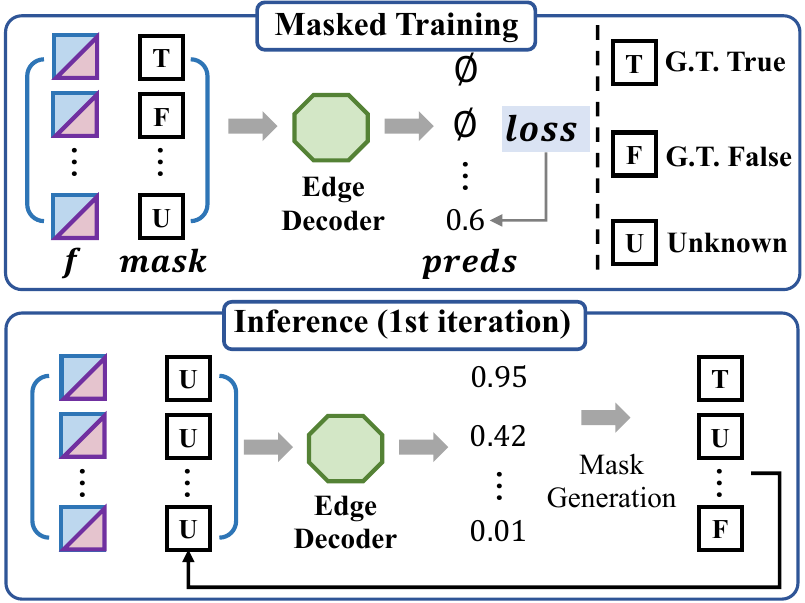}
    \caption{
    (Top) The masked edge learning strategy. (Bottom) The iterative edge inference at the 1st iteration.}
    \label{fig:mask-training}
    \vspace{-1em}
\end{figure}

\subsection{Masked training and iterative inference}
While the proposed architecture works well with vanilla supervised learning, a masked learning strategy inspired by the masked language modeling in BERT~\cite{Devlin2019BERT} can further encourage the learning of structural patterns and boost performance. 
The idea is to randomly provide ground-truth (GT) labels for some edge candidates and only ask to infer missing labels (See Fig.~\ref{fig:mask-training}). 
Each edge candidate has one of the three states: (T) A GT label is given as true; (F) A GT label is given as false; or (U) A GT label is unknown and the network needs to infer. 
We modify the architecture slightly to inject the state information. Concretely, we represent the state as an one-hot encoding vector, convert to 256-dimension by a linear layer, concatenate with $\bm{f}_{\text{coord}}$, and downscale to 256-dimension by another linear layer.

During training, we pick a ratio uniformly from [0, 0.5] and provide GT labels for randomly sampled edge candidates based on the ratio.
At test time, we perform iterative label inference with the image-aware decoder. At the first iteration, the state mask contains all U labels (Fig.~\ref{fig:mask-training}). 
For the second iteration, edges with confidence predictions less than 0.01 (resp. more than 0.9) will have a F (resp. a T) label, while the remaining will have a U label. At the last iteration, a threshold of 0.5 is used to produce the final predictions.
We set the number of iterations to be 3 for all experiments.

\subsection{Corner detector} 
\label{section:corner_detector}

\ourmethod works with any corner detector. We found that an adaptation of HEAT architecture achieves superior corner detection performance and further improves the edge classification.
Concretely, we take the HEAT architecture till the edge-filtering module as a corner detector, as we found that self-attention did not improve corner detection.
Every pixel is a corner candidate and hence becomes a Transformer node. 
The node feature is initialized by the positional encoding of the pixel coordinate, which is half the dimension of the edge feature. Thus we modify $\bm{M}_{\text{coord}}$ to be a  128$\times$256 matrix so that the rest of the architecture remains the same. Non-maximum suppression is applied and we use a threshold of 0.01 to select corner candidates, such that important corners are unlikely to be missed.
The corner model is also trained with a BCE loss. Please see the supplementary for details. 

\section{Experiments}
\label{sec:exp}

\begin{table*}[!t]
    \centering
    \small
    \caption{\textbf{Quantitative evaluations on outdoor architecture reconstruction.} Size: the size of input image. Fully-neural: not using hand-crafted optimization or search techniques. Joint: the edge prediction is trained end-to-end with corner detection.
    The colors \first{cyan} and \second{orange} mark the top-two results with different image sizes. 
    }
    \vspace{-1em}
    {
        \tabcolsep 5pt
	    \begin{tabu}{@{\;} lccc @{\quad\;\;\;\;}  ccccccccc@{\;}}
	        \addlinespace
	        \toprule
	        \multicolumn{4}{l}{Evaluation Type $\rightarrow$} &  \multicolumn{3}{c}{Corner} &  \multicolumn{3}{c}{Edge} &  \multicolumn{3}{c}{Region} \\
	        \cmidrule(lr){5-7} \cmidrule(lr){8-10} \cmidrule(lr){11-13}
	        {Method} & Size & Fully-neural & Joint &  Prec & Recall & F-1 & Prec & Recall & F-1 & Prec & Recall & F-1  \\
	        \midrule
	        IP~\cite{Nauata2020VectorizingWB} & 256 & - & - &  - & - & 74.5 & - & - & 53.1 & - & - & 55.7 \\
	        Exp-Cls~\cite{Zhang2021ExploreClassify} & 256 & - & - &  \first{92.2} & 75.9 & 83.2 & 75.4 & 60.4 & 67.1 & \second{74.9} & 54.7 & \second{63.5} \\
	        ConvMPN~\cite{Zhang2020ConvMPN} & 256 & \cmark & - &  78.0 & 79.7 & 78.8 & 57.0 & 59.7 & 58.1 & 52.4 & 56.5 & 54.4  \\ 
	        HAWP~\cite{Xue2020HAWP} & 256 & \cmark & \cmark &  90.9 & \second{81.2} & \second{85.7} & \second{76.6} & \second{68.1} & \second{72.1} & 74.1 & \second{55.4} & 63.4 \\
	        LETR~\cite{Xu2021LETR} & 256 & \cmark & \cmark & 87.8 & 74.8 & 80.8 & 59.7 & 58.6 & 59.1 & 68.3 & 48.7 & 56.8 \\
	        \ourmethodtable & 256 & \cmark & \cmark & \second{91.7} & \first{83.0} & \first{87.1} & \first{80.6} & \first{72.3} & \first{76.2} & \first{76.4} & \first{65.6} & \first{70.6} \\ 
	        \midrule
	        HAWP~\cite{Xue2020HAWP} & 512 & \cmark & \cmark & \second{90.6} & \second{83.7} & \second{87.0} & \second{78.8} & \second{72.0} & \second{75.2} & \second{77.5} & 57.8 & 66.2 \\
	        LETR~\cite{Xu2021LETR} & 512 & \cmark & \cmark & 90.3 & 79.7 & 84.7 & 64.0 & 71.6 & 67.6 & 77.1 & \second{62.6} & \second{69.1} \\
	        \ourmethodtable & 512 & \cmark & \cmark & \first{90.7} & \first{86.7} & \first{88.7} & \first{82.2} & \first{77.4} & \first{79.7} & \first{79.6} & \first{69.0} & \first{73.9}  \\
    	    \bottomrule
	    \end{tabu}
	}
    \label{tab:outdoor-main}
\end{table*}

We implemented our approach in Python3.7 and Pytorch1.5.1, and used a workstation with a 3.4GHz Xeon and dual NVIDIA RTX 2080 GPUs. 
Our image encoder contains only one Transformer layer while the edge decoders have six. We use ResNet-50 as the backbone to be consistent with the Transformer-based competing method LETR~\cite{Xu2021LETR}. 
The loss balancing weights of the three edge BCE losses are all 1.0, while the weight for corner prediction BCE is 0.05 and 0.1 for the outdoor and indoor tasks, respectively. 
We train our model with Adam optimizer~\cite{Kingma2015AdamAM} with an initial learning rate 2e-4 and a weight decay factor 1e-5. The learning rate decays by a factor of 10 for the last 25$\%$ epochs. With a reference to LETR, our training schedule has 800 epochs for outdoor reconstruction and 400 epochs for floorplan reconstruction, based on the number of training images. Note that the datasets are both small so we set a large epoch number without hyper-parameter search, and the same setting is used for running competing methods. \ourmethod does not apply any post-processing for generating the final planar graph.

\subsection{Competing methods}
We evaluate the proposed approach on two tasks (i.e., outdoor architecture and indoor floorplan reconstructions). For the outdoor task, we compare with
five methods: ConvMPN~\cite{Zhang2020ConvMPN}, IP~\cite{Nauata2020VectorizingWB}, Exp-cls~\cite{Zhang2021ExploreClassify}, HAWP~\cite{Xue2020HAWP} and LETR~\cite{Xu2021LETR}. The first three methods were demonstrated for the same outdoor task, while the latter two methods were for the wire-frame parsing.

\vspace{0.1cm}
\noindent $\bullet$ {\bf ConvMPN} is an improved graph neural network for edge classification and requires a pre-trained corner detector.

\noindent $\bullet$ {\bf IP} and {\bf Exp-cls} rely on heavy optimization or search process to reconstruct a planar graph based on geometric primitives detected by neural networks.

\noindent $\bullet$ {\bf HAWP}~\cite{Xue2020HAWP} is a state-of-the-art wireframe parsing method, which is an improvement over LCNN~\cite{Zhou2019EndtoEndWP} and performs independent edge classification.

\noindent $\bullet$ {\bf LETR} is a Transformer-based line detection framework adapted from the object detection framework DETR~\cite{Carion2020DETR}.

\vspace{0.1cm}

For the indoor task, we compare with four methods: HAWP, LETR, Floor-SP~\cite{Chen2019FloorSPIC}, and MonteFloor~\cite{Stekovic2021MonteFloorEM}. HAWP and LETR have flexible designs with good performance for the outdoor task, and hence are chosen again.

\vspace{0.1cm}
\noindent $\bullet$ {\bf Floor-SP} is one of the state-of-the-art floorplan reconstruction techniques with domain specific system design. The method first uses Mask-RCNN~\cite{He2020MaskR} to obtain room segmentation, and conducts complex optimization procedure.

\noindent $\bullet$ {\bf MonteFloor} is an improvement over Floor-SP with a similar algorithm specifically designed for floorplan reconstruction, where a Monte Carlo tree search algorithm is employed. 

\vspace{0.1cm}
For both outdoor and indoor tasks, we borrow the standard metrics in the literature for the evaluation (See Sect.~\ref{sec:problem}).
We use the numbers reported in the original papers for domain-specific baselines (i.e., ConvMPN, IP, Exp-cls, Floor-SP, and MonteFloor). The publicly available official implementations are used for HAWP and LETR.

\subsection{Quantitative evaluation}

\mypara{Outdoor architecture reconstruction:}
Table~\ref{tab:outdoor-main} presents the main quantitative evaluation.
\ourmethod outperforms all the competing methods on all the F-1 scores, including IP and Exp-Cls, which employ expensive optimization/search methods and are a few orders of magnitude slower than our method.
Note that the outdoor reconstruction task~\cite{Nauata2020VectorizingWB} uses 256$\times$256 images, while HAWP and LETR were demonstrated on larger images in their original papers. Therefore, we also conduct experiments in the 512$\times$512 resolution, where we resize the training/testing images with the same data split.
In the higher resolution, LETR exhibits higher region metrics but poor edge precision compared to HAWP.
Our hypothesis is that LETR relies mostly on the image features and does not learn holistic geometric reasoning over the ``dummy query nodes'', resulting in many false edges and building reconstructions which do not look like buildings. \ourmethod is still a clear winner.

\mypara{Floorplan reconstruction:} 
\begin{table}[!t]
    \centering
    \small
    \caption{\textbf{Quantitative evaluations on floorplan reconstruction.} Results for MonteFloor and Floor-SP are borrowed from MonteFloor paper. The colors \first{cyan} and \second{orange} mark the top-two results.
    }
    \vspace{-1em}
    {
        \tabcolsep 3pt
	    \begin{tabu}{@{\;} lccccccc@{\;}}
	        \addlinespace
	        \toprule
	        \multicolumn{2}{l}{Eval Type $\rightarrow$} & \multicolumn{2}{c}{Room} &  \multicolumn{2}{c}{Corner} &  \multicolumn{2}{c}{Angle} \\
	        \cmidrule(lr){3-4} \cmidrule(lr){5-6} \cmidrule(lr){7-8}
	        {Method} & t(s) & Prec & Recall & Prec & Recall & Prec & Recall \\
	        \midrule
	        HAWP~\cite{Xue2020HAWP} & 0.02 &  0.78 & 0.88 & 0.66 & 0.77 & 0.60 & 0.70  \\ 
	        LETR~\cite{Xu2021LETR} & 0.04 & 0.94 & 0.90 & 0.80 & \second{0.78} & 0.72 & 0.71  \\
	        \ourmethodtable & 0.11 & \first{0.97} & \first{0.94} & \second{0.82} & \first{0.83} & 0.78 & \first{0.79}  \\
	        \hdashline
	        Floor-SP~\cite{Chen2019FloorSPIC} & 785 & 0.89 & 0.88 & {0.81} & 0.73 & \second{0.80} & 0.72 \\
	        MonteFl.~\cite{Stekovic2021MonteFloorEM} & 71 & \second{0.96} & \first{0.94} & \first{0.89} & 0.77 & \first{0.86} & \second{0.75} \\
    	    \bottomrule
	    \end{tabu}
	}
    \label{tab:indoor-main}
\end{table}
Table~\ref{tab:indoor-main} presents the main quantitative evaluation.
The overall performance of \ourmethod surpasses HAWP, LETR, and Floor-SP, and is on par with MonteFloor. It is surprising that our \ourmethod is even comparable to MonteFloor, which utilizes both deep neural networks and Monte Carlo tree search algorithm.
The average inference time of MonteFloor is more than a minute, while our method runs in a dozen milliseconds.

\begin{figure*}[!t]
    \centering
    \includegraphics[width=\textwidth]{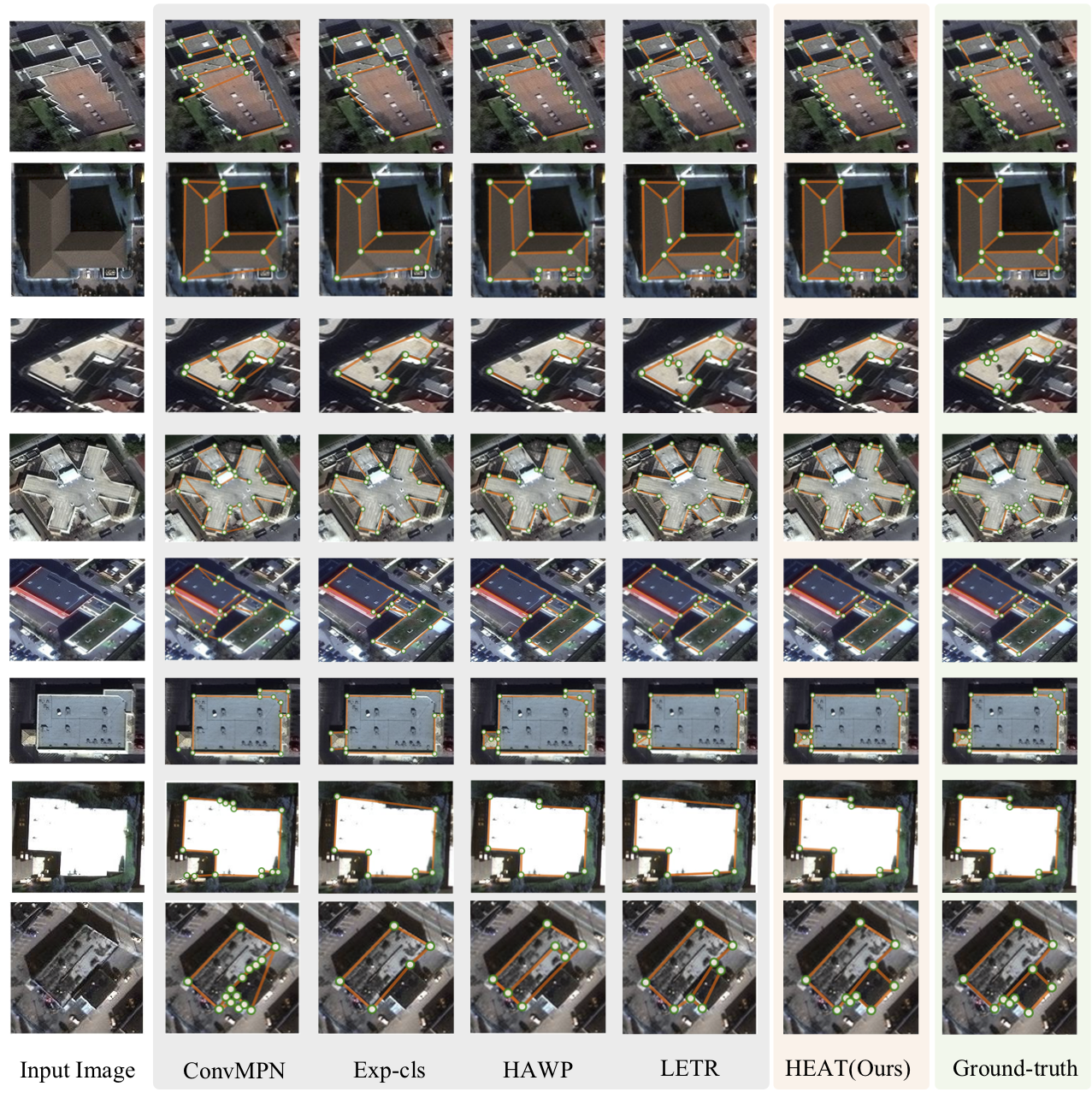}
    \vspace{-2em}
    \caption{\textbf{Qualitative evaluations on outdoor architecture reconstruction} with image size 256$\times$256. Large and complex samples are selected to demonstrate the challenges of the task.
    More results with higher resolution are available in the supplementary material.}
\label{fig:outdoor-qualitative}
\end{figure*}

\subsection{Qualitative evaluation}
\mypara{Outdoor architecture reconstruction:} 
Figure~\ref{fig:outdoor-qualitative} provides qualitative comparisons. The reconstruction quality of \ourmethod is clearly better than the competing methods and close to the ground-truth even on large and complex architectures. Looking carefully at the structures, \ourmethod is especially good at capturing fine details and keep the overall prediction consistent and geometrically valid (\eg, less dangling edges, not disturbed by background buildings). 

\begin{figure}[!t]
    \centering
    \includegraphics[width=\linewidth]{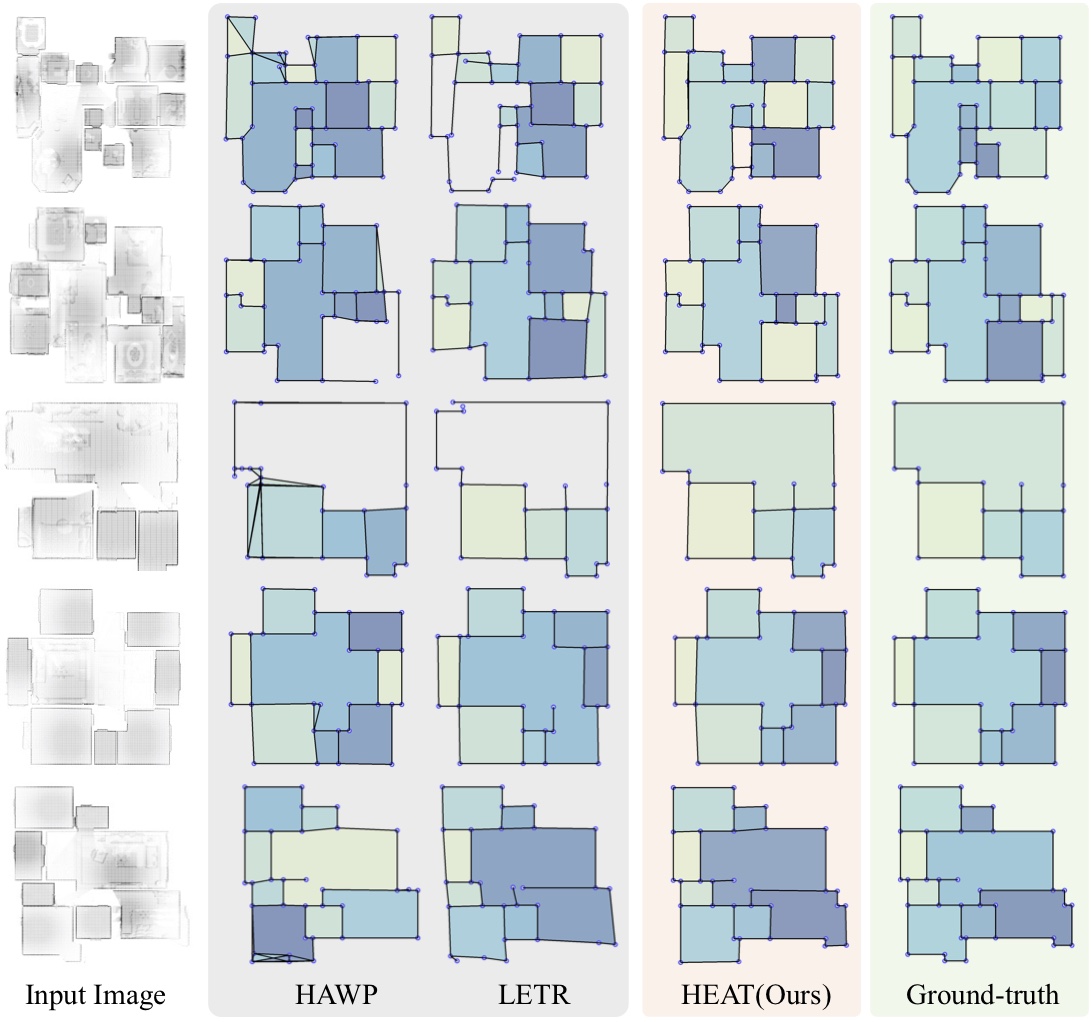}
    \caption{\textbf{Qualitative evaluations on floorplan reconstruction}. Closed polygons in the predicted planar graph are colored while broken ones are not.}
    \label{fig:indoor-qualitative}
\end{figure}

\mypara{Floorplan reconstruction:} 
Figure~\ref{fig:indoor-qualitative} shows the qualitative comparisons.
Only valid polygons of the predicted planar graph are colored. LETR and \ourmethod are better at recovering complete room structures compared to HAWP, as the use of Transformer enables better structured reasoning. \ourmethod still outperforms LETR in terms of producing accurate angles, which is consistent with the numbers in Table~\ref{tab:indoor-main}.

\subsection{Ablation studies}
We conduct extensive ablation studies on the outdoor reconstruction task.

\begin{table}[!t]
    \centering
    \small
    \caption{\textbf{Ablation study for different corner/edge prediction methods}, evaluated with corner/edge/region F-1 scores on outdoor reconstruction. The image size is 256$\times$256. Joint: the corner and edge prediction models are trained end-to-end. FRCNN: pre-trained Faster-RCNN provided by ConvMPN. HG: HourGlass netowork used by HAWP. 
    \vspace{-1em}
    }
    {
        \tabcolsep 3pt
	    \begin{tabu}{@{\;} ccc ccc@{\;}}
	        \addlinespace
	        \toprule
	        Corner Model & Edge Model & Joint & Corner & Edge & Region \\
	        \midrule
	        FRCNN & ConvMPN~\cite{Zhang2020ConvMPN} & - & 78.8 & 58.1 & 54.4  \\ 
	        FRCNN & \ourmethodtable & - & 78.8 & 68.2 & 62.5 \\ 
        \midrule
        HG &  HAWP~\cite{Xue2020HAWP} & \cmark  & 85.7 & 72.1 & 63.4 \\
	    HG &  \ourmethodtable & \cmark & 85.9 & 74.3 & 69.3  \\
        \midrule
        \ourmethodtable & \ourmethodtable & \cmark & 87.1 & 76.2 & 70.6 \\  	\bottomrule
	    \end{tabu}
	}
    \label{tab:corner-edge}
\end{table}

\mypara{Separating corner and edge modules:}
Most methods are combinations of corner detection and edge selection modules. Table~\ref{tab:corner-edge} assesses individual contributions. In particular, we compare three corner models: 
1) a pre-trained Faster-RCNN~\cite{ren2015faster} (FRCNN) used and provided by ConvMPN; 2) an end-to-end trained HourGlass (HG) network used in LCNN and HAWP; and 3) HEAT (ours). The second, the fourth, and the fifth rows share the same HEAT edge model and demonstrate that HEAT corner model makes consistent improvements in all the F1-scores.
Next, we compare three edge selection models: 1) ConvMPN; 2) HAWP; and 3) HEAT (ours).  The first two rows compare ConvMPN and HEAT, while both using FRCNN as the corner model to be fair. Similarly, the next two rows compare HAWP and HEAT, while using HG as the corner model that was used by HAWP. The table again shows that HEAT makes clear improvements over all the competing methods.

\begin{table}[!t]
    \centering
    \small
    \caption{\textbf{Ablation study for the technical components of \ourmethod edge prediction}, evaluated with edge/region F-1 scores of outdoor reconstruction. The image size is 256$\times$256. The pre-trained FRCNN from ConvMPN is used as the corner model.
    }
    \vspace{-1em}
    {
        \tabcolsep 3pt
	    \begin{tabu}{ccccc |cc}
	        \addlinespace
	        \toprule
    	     Coord & D-attn & Dec$^{\text{img}}$ & Mask & Dec$^{\text{geom}}$ & Edge & Region \\
	        \midrule
	        - & \cmark & - & - & - & 59.2 & 21.0  \\
	        \cmark & - & - & - & - & 62.8 & 43.5 \\
	        \cmark & \cmark & - & - & - & 67.3 & 48.6 \\
	        \cmark & \cmark & \cmark & - & - &  67.3 & 60.2 \\
	        \cmark & \cmark & \cmark & \cmark & - & 68.5 & 60.7 \\
	        \cmark & \cmark & \cmark & \cmark & \cmark &  68.2 & 62.5 \\
    	    \bottomrule
	    \end{tabu}
	}
    \label{tab:ablation}
\end{table}

\mypara{\ourmethod system components:}
Table~\ref{tab:ablation} evaluates the five HEAT system components in the edge selection module. The pre-trained FRCNN model by ConvMPN is used for corner detection to be simple. The five columns indicate:

\noindent $\bullet$ [Coord] 
Node initialization with edge coordinates information or not (\ie, a zero vector);

\noindent $\bullet$ [D-attn] The edge deformable attention for image feature extraction or the LoI-Pooling from LCNN~\cite{Zhou2019EndtoEndWP};

\noindent $\bullet$ [Dec$^{\text{img}}$] The image-aware decoder or without it (i.e., using the confidence scores from edge filtering for prediction);

\noindent $\bullet$ [Mask] The masked learning and iterative inference or single-shot training and inference; and

\noindent $\bullet$ [Dec$^{\text{geom}}$] The weight-sharing geometry-only decoder or the image-aware decoder alone.

\vspace{0.1cm}
\noindent
The first row shows that the edge deformable attention cannot work well without proper coordinate features. Other rows in the table validate the contributions of the components, which consistently improve the metrics.

\begin{table}[!t]
    \centering
    \small
    \caption{\textbf{Edge prediction with the geometry-only decoder}, with different types of corners. ``edge filter'' means all remained edge candidates after the edge filtering. ``\ourmethod(geom)'' means using the geometry-only decoder for inference. ``GT'' uses ground-truth information to select corner pairs as the answers, representing the performance upper-bound given the corners. 
    }
    \vspace{-1em}
    {
        \tabcolsep 2pt
	    \begin{tabu}{@{\;} ll cccccc@{\;}}
	        \addlinespace
	        \toprule
	        \multicolumn{2}{c}{Eval Type $\rightarrow$ } & \multicolumn{3}{c}{Edge} &  \multicolumn{3}{c}{Region} \\
	        \cmidrule(lr){3-5}\cmidrule(lr){6-8}
	    Corners   & Method  & Prec & Recall & F-1 & Prec & Recall & F-1  \\
	        \midrule
	         
	        \multirow{4}{*}{FRCNN} & edge filter & 25.8 & 68.5 & 37.5 & 3.4 & 22.4 & 5.9  \\
	         & \ourmethod(geom) & 63.3 & 56.5 & 59.7 & 49.7 & 44.3 & 46.8 \\
	         & \ourmethod & 77.5 & 60.9 & 68.2 & 74.7 & 53.8 & 62.5 \\
	       \cmidrule(lr){2-8}
	         & GT & 83.6 & 66.8 & 74.2 & 80.4 & 63.4 & 70.9 \\
	         \midrule
	         \multirow{4}{*}{GT} & edge filter & 38.5 & 99.9 & 55.6 & 5.8 & 35.9 & 10.0    \\
	         & \ourmethod(geom) & 91.6 & 92.6 & 91.9 & 71.9 & 76.2 & 74.0 \\
	         & \ourmethod & 96.6 & 93.8 & 95.2 & 91.8 & 84.5 & 88.0 \\
	         \cmidrule(lr){2-8}
	         & GT & 100 & 100 & 100 & 100 & 100 & 100 \\
    	    \bottomrule
	    \end{tabu}
	}
    \label{tab:geo_infer}
\end{table}

\mypara{Geometry-only decoder:} Table~\ref{tab:geo_infer} shows the power of the geometry-only decoder that conducts only geometric reasoning without any image information. Concretely, we compare performance of the four edge-selection results: the edge-filtering, the geometry-only decoder, the image-aware decoder (the full \ourmethod), and the ground-truth (GT), while using either FRCNN or GT corners for corner detection. The geometry-only decoder makes significant improvements over the edge-filtering, which is the input of the decoder.
The most striking result is that the geometry-only decoder with the GT corners is much better than the HEAT full system with FRCNN corners. The result demonstrates that HEAT learns to conduct powerful holistic geometric reasoning.

\begin{figure}[!t]
    \centering
    \includegraphics[width=\linewidth]{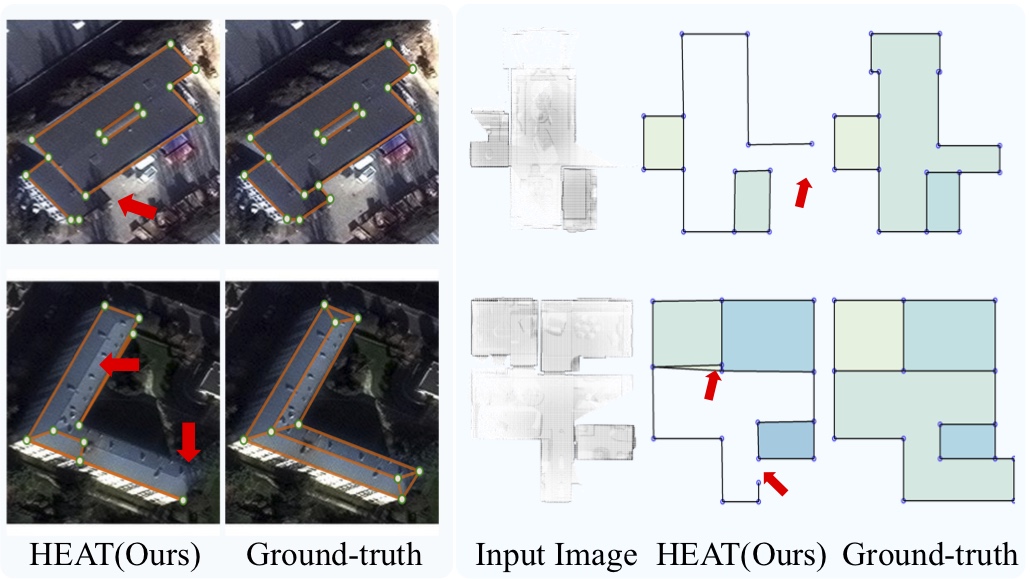}
    \vspace{-1.5em}
    \caption{\textbf{Typical failure modes} on the two reconstruction tasks. }
    \vspace{-1em}
    \label{fig:failure}
\end{figure}

\mypara{Limitations} Figure~\ref{fig:failure} presents failure modes of \ourmethod. The model still misses important corners even with a small threshold, leading to further mistakes in edge prediction.
Rare structures (\eg, the bottom-left L-shape building) are also challenges. We do not touch 3D structured reconstruction (\eg, the 3D wireframe parsing task~\cite{Zhou2019wireframe3D}) with \ourmethod, which can be a potential future work. 

Please see the supplementary for more qualitative visualizations as well as quantitative ablation studies (\eg, details of iterative inference, choice of positional encoding, etc.).

\section{Conclusion}
\label{sec:conclusion}

This paper presents \ourmethod, a novel attention-based neural network that takes 2D raster image and reconstructs a planar graph depicting the underlying geometric structures. \ourmethod detects corners and learns to exploit both image information and geometric patterns between edge candidates in reconstructing a planar graph.
Two weight-sharing Transformer decoders are the technical core of the approach.
The entire system is trained end-to-end with a masked learning strategy and performs iterative inference at test time. Qualitative and quantitative evaluations demonstrate that our method pushes the frontier of end-to-end neural architecture for structured reconstruction.
Extensive ablation studies further justify the design choices.

\mypara{Potential negative societal impact:} Outdoor reconstruction could potentially facilitate satellite imaging for military tasks. Indoor floorplan estimation could raise privacy concerns for house scanning without user permission. 

\mypara{Acknowledgement:} The research is supported by NSERC Discovery Grants, NSERC Discovery Grants Accelerator Supplements, DND/NSERC Discovery Grants, and John R. Evans Leaders Fund (JELF).

{\small
\bibliographystyle{ieee_fullname}
\bibliography{egbib}
}

\appendix

\begin{center}
    \large \textbf{Appendix}
\end{center}

\noindent The appendix is organized as follows:
\begin{itemize}[leftmargin=*,itemsep=1pt]
    \item Sect.\ref{sup:sec:exp}: More experimental results including \begin{itemize}[leftmargin=*,topsep=0pt,itemsep=0pt,noitemsep]
        \item Analysis on the masked learning strategy, the geometry-only edge decoder, and the iterative inference;
        \item Ablation studies on the choice of positional encoding for representing edge coordinates;
        \item Qualitative evaluation for more testing samples on the two structured reconstruction benchmarks (See
        \href{https://drive.google.com/file/d/1L2Gr35aFj8VOmpGZBVVozt4zMaFE8pWF/view?usp=sharing}{outdoor\_qualitative.pdf} and  \href{https://drive.google.com/file/d/16tB-e59MKc3sDi4larEm1co3PoHT-Mck/view?usp=sharing}{indoor\_qualitative.pdf}).
    \end{itemize}
    \item Sect~\ref{sup:sec-corner}: Details of our corner detection module adapted from the HEAT edge classification architecture.
    \item Sect.\ref{sup:sec:impl}: Additional implementation details such as the choice of hyper-parameters, training setups for different experiments, training data preparation, and 
    how we reproduced the competing methods.
\end{itemize}

\section{Additional Experimental Results}
\label{sup:sec:exp}

\mypara{Masked learning, geometry-only decoder, and iterative inference:}
Table~\ref{tab:mask_and_twob} complements the Table 4 of the main paper by providing more details about the masked learning strategy the geometry-only decoder, and the iterative inference.. The masked learning strategy alone marginally improves the region scores while adding the geometry-only decoder clearly boosts the region-level performance. These results suggest that the image-aware decoder might over-fit to the image features and neglect the geometric patterns revealed by the coordinate features. The geometry-only decoder could effectively alleviate the above issue as a regularization by sharing weights with the image-aware decoder and conducting geometry-only inference.

\mypara{Choice of positional encoding:}
Table~\ref{tab:coord_enc} provides an ablation study for the choice of positional encoding for edge coordinates. The first row shows that a proper positional encoding is vital for region-level performance. Besides, the learnable embedding is worse than the trigonometric encoding, potentially because the trigonometric positional encoding preserves useful ordinal priors (\eg, relative distance) that are hard to be learned automatically from data. 

\begin{table}[t]
    \centering
    \small
    \caption{Detailed ablation study for masked learning strategy, iterative inference, and the geometry-only (geom-only) decoder. ``Iter'' denotes the number of inference iterations. The pre-trained Faster-RCNN from ConvMPN~\cite{Zhang2020ConvMPN} is used for corner detection.}
    {
        \tabcolsep 3pt
	    \begin{tabu}{@{\;} ccc  @{\quad\;\;} cccccc@{\;}}
	        \addlinespace
	        \toprule
	        \multicolumn{3}{l}{{Eval Type $\rightarrow$}} &  \multicolumn{3}{c}{Edge} &  \multicolumn{3}{c}{Region} \\
	        \cmidrule(lr){4-6}\cmidrule(lr){7-9}
	       Mask & Dec$^\text{geom}$ & Iter  & Prec & Recall & F-1 & Prec & Recall & F-1  \\
	        \midrule
	        - & - & 1 & 75.7 & 60.5 & 67.3 & 74.1 & 50.7 & 60.2 \\
	        \midrule
	        \cmark & - & 1 & 77.4 & 61.2 & 68.3 & 76.4 & 49.7 & 60.2 \\
	        \cmark & - & 2 & 77.8 & 61.0 & 68.4 & 75.3 & 49.4 & 59.7\\
	        \cmark & - & 3 & 77.9 & 61.2 & 68.5 & 76.0 & 50.6 & 60.7\\
	        \midrule
	        \cmark & \cmark & 1 & 76.7 & 60.4 & 67.6 & 73.7 & 52.2 & 61.1 \\
	        \cmark & \cmark & 2 & 77.5 & 60.8 & 68.1 & 75.0 & 53.6 & 62.5 \\
	        \cmark & \cmark & 3 & 77.5 & 60.9 & 68.2 & 74.7 & 53.8 & 62.5 \\
    	    \bottomrule
	    \end{tabu}
	}
    \label{tab:mask_and_twob}
\end{table}
\begin{table}[t]
    \centering
    \small
    \caption{Ablation study on the choice of coordinate encoding. ``Learn'' means using a learnable embedding for each discrete value. ``Sin/Cos'' is the trigonometric positional encoding used by \ourmethod.
    }
    {
        \tabcolsep 3pt
	    \begin{tabu}{@{\;} l@{\quad\;\;} cccccc@{\;}}
	        \addlinespace
	        \toprule
	        {Eval Type $\rightarrow$ } & \multicolumn{3}{c}{Edge} &  \multicolumn{3}{c}{Region} \\
	        \cmidrule(lr){2-4}\cmidrule(lr){5-7}
	        Coord Enc.  & Prec & Recall & F-1 & Prec & Recall & F-1  \\
	        \midrule
	        $\varnothing$ & 67.2 & 58.9 & 62.7 & 29.7 & 39.6 & 33.9 \\
	        Learn &  76.2 & 61.2 & 67.9 & 75.3 & 49.7 & 60.7 \\
	        Sin/Cos & 77.5 & 60.9 & 68.2 & 74.7 & 53.8 & 62.5  \\
    	    \bottomrule
	    \end{tabu}
	}
    \label{tab:coord_enc}
\end{table}

\mypara{More qualitative results}
The files \href{https://drive.google.com/file/d/1L2Gr35aFj8VOmpGZBVVozt4zMaFE8pWF/view?usp=sharing}{outdoor\_qualitative.pdf} and  \href{https://drive.google.com/file/d/16tB-e59MKc3sDi4larEm1co3PoHT-Mck/view?usp=sharing}{indoor\_qualitative.pdf} provide additional high-resolution qualitative results for outdoor architecture reconstruction and floorplan reconstruction, respectively. The presentation formats are the same as the qualitative figures in the main paper. Due to the size limit, we randomly pick 100 testing samples for each task. Please enlarge the figures to assess the details.


\section{\ourmethod Corner Detection}
\label{sup:sec-corner}
This section explains the details of the corner detector, which is an adaptation of our \ourmethod edge classification architecture. The \ourmethod-based corner detection and edge classification modules are trained end-to-end.

\begin{figure}[ht]
\centering
\includegraphics[width=\linewidth]{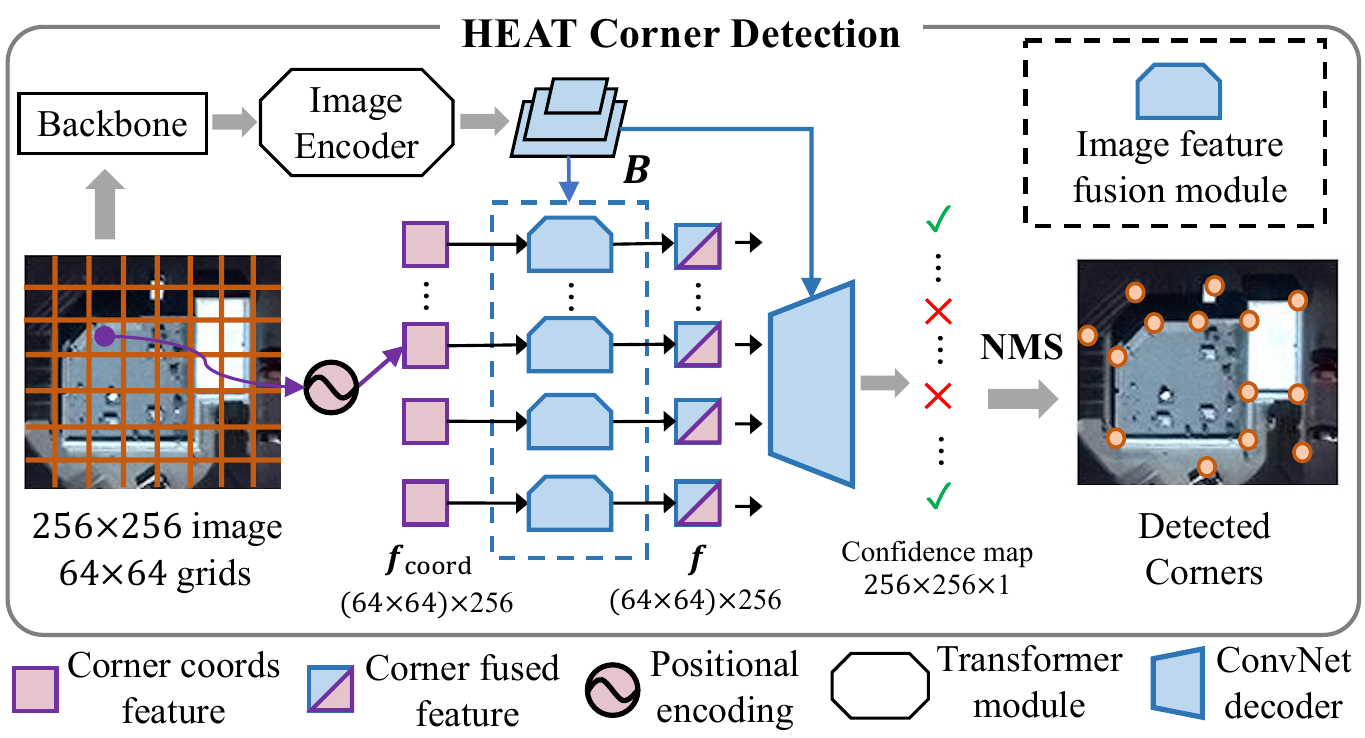}
\caption{\textbf{The corner detection model adapted from \ourmethod architecture}, illustrated with image size 256$\times$256. We take the \ourmethod architecture (Fig.2a of the main paper) until the edge-filtering module as the corner detector. Each node is a 4$\times$4 super-pixel. A ConvNet decoder takes the node features and produces a 256$\times$256 confidence map, Non-maximum suppression (NMS) is applied to the confidence map to produce the final corner detection results.} 
\label{fig:corner_heat}
\end{figure}
Figure~\ref{fig:corner_heat} illustrates our corner detection model adapted from \ourmethod architecture. Pixels are the corner candidates and thus become the nodes. Instead of making every pixel in the 256$\times$256 image space as a candidate, we make each 4$\times$4 super-pixel as a node to reduce the memory cost. An extra MLP takes the concatenated coordinate features of the 16 pixels inside a super-pixel and produces the $\bm{f}_{\text{coord}}$ for each node. After the image feature fusion, a ConvNet decoder converts the 64$\times$64$\times$256 feature maps into the final 256$\times$256 confidence map. The ConvNet decoder consists of stacks of convolution layers and up-sampling layers, as well as a final linear layer for producing the confidence. Non-maximum suppression is applied to the confidence map to produce the final corner detection results.

We train the above corner detection model jointly with \ourmethod edge classification. The corner and edge models share the same ResNet backbone. The training data for the edge model are generated on the fly based on the corner detection results. See Sect.~\ref{sup:sec:impl} for details about training data preparation.

\section{Additional Implementation Details}
\label{sup:sec:impl}

\mypara{Hyper-parameter and training settings. }
As mentioned in the main paper, there are four binary cross-entropy (BCE) losses in the full \ourmethod framework: one for corner and three for edge.  We use a weight of 3.0 (resp. 10.0) for positive samples to balance the positive and negative samples for the edge (resp. corner) BCE loss. 

We apply non-maximum suppression (NMS) to the \ourmethod corner prediction results to clean up the detected corners. Non-maximum predictions inside a local 5$\times$5 window are suppressed.  
When using a pre-trained corner detection model (\ie, the Faster-RCNN provided by ConvMPN~\cite{Zhang2020ConvMPN}) in our ablation studies on the outdoor reconstruction benchmark, the number of training epochs is reduced from 800 to 500 as we only need to train the edge classification part alone, and the corner BCE is discarded. All other training settings are exactly the same as the full \ourmethod. 
For all experiments, we simply take the checkpoint from the last training epoch for evaluation.

\mypara{Training data preparation.} We conduct random flipping and random rotation for data augmentation when training the models on the outdoor reconstruction task. However, only random flipping is applied for indoor reconstruction since random rotation always makes the planar graph surpass the image boundary. 

For producing corner labels, we first produce a label map with the same resolution as the input image, and then apply a Gaussian blur (with sigma=2) to the label map to alleviate the class imbalance.   

For producing edge labels, we follow three steps: 1) Match ground-truth corners with detected corners. A detected corner and a ground-truth corner are matched if their distance is smaller than 5 pixels and both of them are not matched with other corners; 2) Generate training-time corner candidates. We produce the set of corner candidates by merging all the detected corners and ground-truth corners, but removing the ground-truth corners that are matched by detected corners; 3) Enumerate corner pairs and assign edge labels. We enumerate all pairs of corner candidates to generate the edge candidates. The label of an edge candidate is true if and only if each of its endpoints is either a ground-truth corner or a matched detected corner, otherwise the label is false.

\mypara{Running competing approaches.}
We explain how we run the competing approaches to evaluate them on the two structured reconstruction benchmarks: 

\noindent $\bullet$ {HAWP~\cite{Xue2020HAWP}:}  We adapt the official HAWP implementation\footnote{\url{https://github.com/cherubicXN/hawp}} for the two structured reconstruction tasks with two modifications: 1) we change the image resolution according to the experimental setups and 2) we increase the number of training epochs to make it the same as \ourmethod and LETR. We found that increasing the number of training epochs does not improve HAWP on its original wireframe parsing task, but can clearly improve its performance on the structured reconstruction tasks.

\noindent $\bullet$ {LETR~\cite{Xu2021LETR}:} We adapt the official implementation of LETR\footnote{\url{https://github.com/mlpc-ucsd/LETR}} for the two structured reconstruction tasks with three modifications: 1) we change the image resolution based on the experimental setups; 2) we search for the best hyper-parameter for the number of ``dummy query nodes'' and change it from 1000 to 100, which is also consistent with the dataset stats in Sect.3 of the main paper (the model with default settings cannot converge on our two benchmarks); and 3) we run a simple post-processing to merge neighbouring corners within 10 pixels to get a cleaner planar graph. Note that we strictly follow LETR's three-stage training pipeline.

\noindent $\bullet$ {ConvMPN~\cite{Zhang2020ConvMPN} and Exp-cls~\cite{Zhang2021ExploreClassify}:} We run the released official checkpoints\footnote{\url{https://github.com/zhangfuyang/Conv-MPN} and \url{https://github.com/zhangfuyang/search_evaluate}} to get the quantitative evaluation results and corresponding qualitative visualizations.  

\noindent $\bullet$ {Others:} For other domain-specific approaches (\ie, IP~\cite{Nauata2020VectorizingWB}, MonteFloor~\cite{Stekovic2021MonteFloorEM}, Floor-SP~\cite{Chen2019FloorSPIC}), we directly borrow their evaluation results from previous papers.

\end{document}